\documentclass[journal]{IEEEtran}
\ifCLASSINFOpdf
  \usepackage[pdftex]{graphicx}
\else
  \usepackage[dvips]{graphicx}
\fi
\begin{document}
\title{Learning Hidden Markov Models using Non-Negative Matrix Factorization}

\newcommand{\one}{\ensuremath{\mathbf{1}}}

\author{George~Cybenko,~\IEEEmembership{Fellow,~IEEE,}
                and~Valentino~Crespi,~\IEEEmembership{Member,~IEEE}% <-this % stops a space
\thanks{G. Cybenko is with the Thayer School of Engineering, Dartmouth College,
Hanover, NH 03755 USA e-mail: gvc@dartmouth.edu.}% <-this % stops a space
\thanks{V. Crespi is with the Department of Computer Science, California State University at
Los Angeles, LA, 90032 USA e-mail: vcrespi@calstatela.edu.} % stops a space
\thanks{Manuscript submitted September 2008}}


\markboth{Submitted to IEEE Transactions on Information Theory, September~2008}%
{Shell \MakeLowercase{\textit{et al.}}: Bare Demo of IEEEtran.cls for Journals}



\maketitle

\begin{abstract}
%\boldmath
The Baum-Welch algorithm together with its derivatives and
variations has been the main technique for learning Hidden Markov
Models (HMM) from observational data.  We present an HMM learning
algorithm based on the non-negative matrix factorization (NMF) of
higher order Markovian statistics that is structurally different
from the Baum-Welch and its associated approaches. The described
algorithm supports estimation of the number of recurrent states of
an HMM and iterates the non-negative matrix factorization (NMF)
algorithm to improve the learned HMM parameters.   Numerical
examples are provided as well.
\end{abstract}



% \fi

\IEEEpeerreviewmaketitle

\section{Introduction}


Hidden Markov Models (HMM) have been successfully used to model
stochastic systems arising in a variety of applications ranging from
biology to engineering to
finance~\cite{Rabiner:1989,SWT:2001,AB:2000,RST:1994,PDR:2005,KBSSH:1993}.
Following accepted notation for representing the parameters and
structure of HMM's (see
\cite{finesso:2006,VTHCC-1:2005,VTHCC-2:2005,Rabiner:1989,Ephraim:2002}
for example), we will use the following terminology and definitions:
\begin{enumerate}
\item
$N$ is the number of states of the Markov chain underlying the
HMM. The state space is ${\cal S} = \{S_1, ... , S_N \}$ and the system's
state process at time $t$ is denoted by $x_t$;
\item $M$ is the number of distinct observables or symbols generated
  by the HMM.  The set of possible observables is ${\cal V}= \{v_1
  ,..., v_M\}$ and the observation process at time $t$ is denoted by
  $y_t$. We denote by $y_{t_1}^{t_2}$ the subprocess
  $y_{t_1}y_{t_1+1}\ldots y_{t_2}$;
\item
The joint probabilities
$$a_{ij}(k) = P(x_{t+1} = S_j,y_{t+1}=v_k | x_t = S_i); $$
are the time-invariant probabilities of transitioning to state $S_j$
at time $t+1$ and emitting observation $v_k$ given that at time $t$
the system was in state $S_i$. Observation $v_k$ is emitted during the
transition from state $S_i$ to state $S_j$. We use $A(k)=(a_{ij}(k))$
to denote the matrix of state transition probabilities that emit the
same symbol $v_k$. Note that $A=\sum_{k}A(k)$ is the stochastic matrix
representing the Markov chain state process $x_t$.
\item
The initial state distribution, at time $t=1$, is given by $\Gamma
= \{\gamma_1 ,...,  \gamma_N\}$ where $\gamma_i = P(x_1 = S_i)
\geq 0$ and $\sum_i \gamma_i = 1$.
\end{enumerate}
Collectively, matrices $A(k)$ and $\Gamma$ completely define the
HMM and we say that a model for the HMM is $\lambda = ( \{A(k)\ |\
1\leq k\leq M\}, \Gamma )$.

We present an algorithm for {\em learning} an HMM from single or
multiple observation sequences.  The traditional approach for
learning an HMM is the Baum-Welch Algorithm~\cite{Rabiner:1989}
which has been extended in a variety of ways by others
% REF: ihmm?
\cite{willsky:2008,blei04variational,teh03hierarchical}.

Recently, a novel and promising approach to the HMM approximation
  problem was proposed by Finesso et al.~\cite{finesso:2008}. That
  approach is based on Anderson's HMM stochastic realization
  technique~\cite{anderson:1999} which demonstrates that a positive
  factorization of a certain Hankel matrix (consisting of observation
  string probabilities) can be used to recover the hidden Markov
  model's probability matrices. Finesso and his coauthors used
  recently developed non-negative matrix factorization (NMF)
  algorithms~\cite{Seung:2000} to express those stochastic realization
  techniques as an operational algorithm. Earlier ideas in that vein
  were anticipated by Upper in 1997~\cite{upper:1997}, although that
  work did not benefit from HMM stochastic realization techniques or
  NMF algorithms, both of which were developed after 1997.

  Methods based on stochastic realization techniques, including the
  one presented here, are fundamentally different from Baum-Welch
  based methods in that the algorithms use as input observation
  sequence {\em probabilities} as opposed to raw observation {\em
    sequences}. Anderson's and Finesso's approaches use system
  realization methods while our algorithm is in the spirit of the
  Myhill-Nerode~\cite{myhill_nerode} construction for building
  automata from languages. In the Myhill-Nerode construction, states
  are defined as equivalence classes of pasts which produce the same
  futures. In an HMM, the ``future'' of a state is a probability
  distribution over future observations.  Following this intuition we
  derive our result in a manner that appears comparatively more
  concise and elementary, in relation to the aforementioned approaches
  by Anderson and Finesso.

At a conceptual level, our algorithm operates as follows. We first
estimate the matrix of an observation sequence's high order
statistics.  This matrix has a natural non-negative matrix
factorization (NMF) ~\cite{Seung:2000} which can be interpreted in
terms of the probability distribution of future observations given
the current state of the underlying Markov Chain.  Once estimated,
these probability distributions can be used to directly estimate
the transition probabilities of the HMM.

The estimated HMM parameters can be used, in turn, to compute the
NMF matrix factors as well as the underlying higher order
correlation matrix from data generated by the estimated HMM. We
present a simple example in which an NMF factorization is exact
but does not correspond to any HMM.  This is a fact that can be
established by comparing the factors  computed by the NMF with the
factors computed by the estimated HMM parameters.  This kind of
comparison is not possible with other approaches
\cite{finesso:2008}.

It is important to point out that the optimal non-negative matrix
factorization of a positive matrix is known to be NP-Hard in the
general case ~\cite{vavasis:2007}, so in practice one computes
only locally optimal factorizations. As we will show through
examples, the repeated iteration of the factorization and
transition probability estimation steps improves the
factorizations and overall model estimation. Details are provided
below.

\subsection{Preliminaries and Notation}
The only input to our algorithm is an observation sequence of
length $T$ of the HMM, namely:
$${\cal O}_{1:T} = {\cal O}_1 {\cal O}_2 ... {\cal O}_T$$
where ${\cal O}_t \in {\cal V}$ is the HMM output at observation time
$t$.

We do not assume that the observation time $t=1$ coincides with the
process' initial state so that the initial distribution of states is
not necessarily governed by $\Gamma$.  In fact, at present, our
algorithm is capable of learning only the ergodic partition of an HMM,
namely the set of states that are recurrent.  Consequently, our model
of an HMM refers only to the transition probability component
$\lambda=\{A(k)\}_k$ that identifies this ergodic partition (see
\cite{berman:1994,billingsley:1995} for some background on this
concept).

Given ${\cal O}_{1:T}$, we construct two summary statistics
represented as matrices $R^{p,s}$ and $F^{p,s}$ for positive
integers $p$ and $s$. $R^{p,s}$ is simply a histogram of
contiguous prefix-suffix combinations whose rows are indexed by
observations subsequences of length $p$ and columns are indexed by
observation subsequences of length $s$.

If there are $M$ symbols in the observation alphabet, then
$R^{p,s}$ is an $M^p$ by $M^s$ matrix whose $(i,j)$th entry is the
number of times the prefix substring corresponding to $i$ is
immediately followed by the substring corresponding to $j$. The
correspondence between strings and integers is lexicographic in
our examples below although any other correspondence will do as
well.

The matrix $F^{p,s}$ is simply $R^{p,s}$ normalized to be row
stochastic.  Specifically, if $G = (g_i)$ where $g_i = \sum_j
R_{i,j}^{p,s}$ then $F_{i,j}^{p,s} = R_{i,j}^{p,s}/g_i$ for $g_i
\neq 0$ and $F_{i,j}^{p,s} = 0$ for $g_i = 0$. Rows of $R^{p,s}$,
and correspondingly $F^{p,s}$, are zero if the prefix
corresponding to the row label is not observed in the data. Zero
rows of these matrices can be deleted reducing the size of the
matrices without affecting the algorithm describe below.
Accordingly, $F^{p,s}$ is constructed to be row stochastic.

Entry $F_{u,v}^{p,s}$ is essentially an estimate of $P(V|U)$ the
probability of observing observation sequence $V$ of length $s$,
indexed by $v$, following observation sequence $U$ of length $p$,
indexed by $u$ (see the work of Marton, Katalin and
  Shields~\cite{marton:1994} for a study of the accuracy of such
  estimates).

Note that while $R$, $F$ and $G$ have exponentially many rows and
columns with respect to $p$ and $s$, the actual number of nonzero
entries in these matrices are bounded above by $T$ so that, stored
as sparse matrices, they require no more storage than the original
observation sequence.  Note that Baum-Welch methods require
storing and repeatedly accessing the original observation
sequence.

A simple but key observation about states of an HMM is that each state
of an HMM induces a probability distribution on symbol subsequences of
any length $s$.  Specifically, suppose an HMM, $\lambda$, is in state
$S_{i_0}$ (having not yet emitted an observation in that state) and
consider the symbol subsequence $V = v_{j_1} v_{j_2} ... v_{j_s}$.
Then
$$P(V|S_{i_0},s,\lambda) = P( y_{t+1}^{t+s} = v_{j_1} v_{j_2} ... v_{j_s} | x_{t}=S_{i_0})$$
is independent of $t$ under the ergodic assumption and can be computed from the $A(k)$'s according to
\begin{equation}\label{pV}
P(V|S_{i_0},s,\lambda) = e_{S_{i_0}}'\prod_{r=1}^s
A(j_r)\one\ ,
\end{equation}
where $e_i$ denotes the $(0,1)$-vector whose only nonzero entry is
in position $i$ and $\one=[1\ 1\ \ldots\ 1]'$.  Call this
probability distribution on substrings of length $s$,
$P(\cdot|S_i,s, \lambda)$. It is known that the distributions
$P(\cdot|S_i,s, \lambda)$ for $p+s \geq 2N-1$ are complete
characterizations of the ergodic states of the HMM with respect to
the observables of the HMM \cite{carlyle:1969,finesso:2008}.

We now focus attention on substrings that {\em precede} state
occupancy in the HMM's underlying Markov chain.  Over the course
of a long observation sequence such as ${\cal O}_{1:T}$, there is
some probability, $P(S_i|U,p,\lambda)$ that the HMM is in state
$i$ given that we have just observed the length $p$ substring
$U=v_{j_1}v_{j_2}\ldots v_{j_p}$. These probabilities can be
computed from the $A(k)$'s according to
\begin{equation}\label{pS}
P(S_{i_0}|U,p,\lambda) = \frac{\pi'\prod_{r=1}^p
A(j_r)e_{s_{i_0}}}{P(U|p,\lambda)}\ ,
\end{equation}
where $\pi$ is the stationary distribution of the underlying
Markov chain process and $P(U|p,\lambda)=\pi'\prod_{r=1}^p
A(j_r)\one$.

Note that formulas (1) and (2) are closely related to
 computations arising in the classical Viterbi algorithm \cite{Rabiner:1989}.

  Let $U,V$ be two strings of
observations of length $p$ and $s$ respectively. Let $U$ and $V$
be identified with integers $u$ and $v$ as already explained
before so that $P(V|U,\lambda)=F_{u,v}^{p,s}$. Assume $V$ was
emitted after time $t$ and $U$ immediately preceded $V$. We call
$U$ the prefix string and $V$ the suffix string. Then by applying
elementary properties of probability we can write:
\begin{eqnarray*}
  F_{u,v}^{p,s} & \sim & P(y_{t+1}^{t+s}=V\ |\ y_{t-p+1}^t=U,\lambda)\\
  & = & \sum_{k=1}^N P(y_{t+1}^{t+s}=V,x_t=S_k|y_{t-p+1}^t=U,\lambda)\\
%  & = & \sum_{k=1}^NP(V|S_k,U,\lambda)P(S_k|U,\lambda)\\
  & = & \sum_{k=1}^NP(V|S_k,s,\lambda)P(S_k|U,p,\lambda)\ .
\end{eqnarray*}
Consequently we can express the distribution $F_{u,:}^{p,s}\sim
P(\cdot|U,\lambda)$ as a {\em mixture}
\begin{equation}\label{F}
F_{u,:}^{p,s}\sim\sum_{k=1}^NP(\cdot|S_k,s,\lambda)P(S_k|U,p,\lambda)\
.
\end{equation}
If the underlying state process $x_t$ is ergodic then in the limit as
$T\rightarrow\infty$ relation (\ref{F}) becomes an equality {\em
  almost surely}. As a result of the above observations, for
sufficiently large $p$ and $s$, the matrix $F^{p,s}$ has the following
properties:
\begin{itemize}
\item
${\rm rank}(F^{p,s}) \leq N$,  where $N$ is the minimal number of
states representing the HMM, $\lambda$;
\item
Each row of $F^{p,s}$ is a convex combination (mixture) of the $N$
generators, $P(\cdot|S_i,s, \lambda)$, for $i=1,2,\ldots,N$.;
\item Letting $D$ be the $N\times M^s$ nonnegative matrix whose rows are
  the distributions $P(\cdot|S_k,s,\lambda)$, i.e., $D_{k,:}=
  P(\cdot|S_k,s,\lambda)$, for $k=1,2,\ldots,N$, we can
  rewrite~(\ref{F}) as
  $$
  F_{u,:}^{p,s}\sim [P(S_1|U,p,\lambda)\ P(S_2|U,p,\lambda)\ \cdots\
  P(S_N|U,p,\lambda)]* D\ .
  $$
  Consequently, if we let $C=(c_{u,k})$ be the $M^p\times N$ nonnegative matrix
  with $c_{u,k}=P(S_k|U,p,\lambda)$ we can write $F^{p,s}\sim
  C*D$. Observe that $C$ and $D$ are both (row) stochastic.

\item The factorization depends on the model $\lambda$. Moreover
  factors $C$ and $D$ can be computed directly from $\lambda$ using
  ~(\ref{pV}) and~(\ref{pS}). Consequently, the size of the smallest
  model compatible with the data is equal to prank$(F^{p,s})$, the
  {\em positive rank} of $F^{p,s}$. (The positive rank, prank($A$),
  of an $m\times n$ nonnegative matrix $A$ is the smallest integer $N$
  such that $A$ factors in the product of two nonnegative matrices of
  dimensions $m\times N$ and $N\times n$ respectively.)  It is known
  that ${\rm rank}(A)\leq {\rm prank}(A)\leq\min\{m,n\}$ and that the
  computation of prank$(A)$ is
  NP-hard~\cite{vavasis:2007,vandenhof:1999}. So it would appear that
  in general it is NP-hard to estimate $N$ given $F^{p,s}$ even in
  ideal conditions ($T\rightarrow\infty$) since ${\rm
    rank}(F^{p,s})\leq N$.  However, it is not obvious how difficult
  it is to estimate when ${\rm prank}(F^{p,s})<{\rm rank}(F^{p,s})$ in
  the case $F^{p,s}$ was built from a typical realization of an
  HMM. In fact, typically ${\rm rank}([P(V|U)]_{U,V})\leq {\rm
    rank}(P(V|S)_{S,V})\leq {\rm prank}([P(V|U)]_{U,V})$ but, in
  ``noisy'' conditions, we observe ${\rm rank}(D)\leq {\rm
    prank}(F^{p,s})<{\rm rank}(F^{p,s})$.  We discuss an example at
  the end of this paper that illustrates the open problems and
  challenges. One way to circumvent the problem of guessing $N$ is to
  apply statistical methods directly to the observation sequence,
  without building any intermediate models as done in~\cite{liu:1994}.
\end{itemize}
To summarize this discussion, note that the matrix $F^{p,s}$ is based
on the distribution of length $p$ prefixes and corresponding length
$s$ suffixes and completely characterizes an HMM providing
$M^p\geq N,s\geq 2N-1$. Its positive rank is, in ideal
conditions, equal to the minimal number of states in the underlying
Markov chain. Moreover, an appropriately constructed factorization of
$F^{p,s}$ exposes the state transition and emission probabilities of
the HMM. It is well known that any two $N$-state HMMs consistent
  with the same conditional statistics $[P(V|S)]_{S\in{\cal S},V\in
    {\cal V}^{2N-1}}$ generate the same finite dimensional
    distributions and so are, in this sense,
    equivalent~\cite{paz:1971}. The algorithm presented below
  extracts the state transition matrices, $\{A(k)\}_k$ from this
  factorization. In turn, as shown above, the $A(k)$'s can be used to
  construct the probability distributions over suffixes that generate
  $F^{p,s}$ and so can be used to compute a new factorization.  This
  iteration is essentially the basis for our algorithm.

In the machine learning context, we have access only to a finite
amount of observation data ($T$ bounded). Consequently
rank$(F^{p,s})$ will be generally higher than $N$. This requires a
decision about the HMM's order, $N$, not unlike that arising in
principal component analysis (PCA) \cite{comon:1994} to estimate
the number of components.

\section{The Algorithm}
Based on the above discussion, our algorithm is outlined below.
Numerical examples with discussions follow the formal description.

\begin{enumerate}

\item Compute $F^{p,s}$ and $G$ from the input observation
data, ${\cal O}_{1:T}$, defined above.

\item Estimate the number of states, $N$, by analyzing the
 either
  $F^{p,s}$ or ${\rm diag}(G)*F^{p,s}$, both computed in Step 1. In
  the cases in which ${\rm prank}(F^{p,s})={\rm rank}(F^{p,s})$
  (e.g. when ${\rm rank}(F)\leq 2$) one typical way to obtain this
    estimate is to compute the SVD (singular value decomposition) of
    the aforementioned matrices and then observe the rate of decrease
    of the singular values. For $T$ sufficiently large a significant
    gap between the $N^{\rm th}$ and the $(N+1)^{\rm th}$ largest
    singular value becomes appreciable.  Note that since prank
    $\geq$ rank, an estimate based on the singular values is a
    lower bound for the order of the HMM.

  \item Estimate distributions $P(\cdot|S_i,s,\lambda)$, for
    $i=1,2,\ldots,N$. This step is achieved through the
    Nonnegative Matrix Factorization (NMF) of $F^{p,s}$. This
    yields
    $F^{p,s}\approx C*D$ with $D_{i,:}\approx P(\cdot|S_i,s,\lambda)$
    as observed before.

  Note that because of the finiteness of $T$ in general
  prank$(F^{p,s})>N$. So it is necessary to solve the {\em
    approximate} NMF which consists of determining $C$ and $D$ of
  dimensions $M^p\times N$ and $N\times M^s$ respectively that
  minimize $D_{ID}(F^{p,s}||C*D)$, where
  $$D_{ID}(K||W)=\sum_{ij}(K_{ij}\log
  \frac{K_{ij}}{W_{ij}}-K_{ij}+W_{ij})$$ is the I-divergence
  function~\cite{finesso:2006} (observe that if $\one' K\one = \one'
  W\one=1$ then $D_{ID}(K||W)=\sum_{i,j}K_{i,j}\log K_{i,j}/W_{i,j}$
  so the I-divergence function is a generalization of the
  Kullback-Leibler distance between probability distributions). This
  optimization problem can be solved through iterative
  methods~\cite{Seung:2000,jcohen:1993} that require initial matrices
  $C_0,D_0$ and can only be guaranteed to converge to local
  optima. After executing this step, we have a locally optimal
  estimate of the true distributions $P(\cdot|S_i,s,\lambda)$.

\item Estimate matrices $A(k)$, $k=1,2,\ldots,N$, from $D$. Let us
  consider $A(1)=(a_{i,j}(1))$, the other matrices are estimated in a
  similar manner. Let $V^{(s-1)}=v_{j_1}v_{j_2}\cdots v_{j_{s-1}}$ be
  a generic sequence of $s-1$ observations. Then by marginalization we
  can write
  \[
  P(V^{(s-1)}|S_i,s-1,\lambda)=\sum_{k=1}^M P(V^{(s-1)}v_k|S_i,s,\lambda)\ .
  \]
  Consequently, the conditional distributions over suffixes of length
  $s-1$, $P(\cdot|S_i,s-1,\lambda)$, can be estimated from $D$ by
  adding columns of $D$ appropriately. Let $H$ be the matrix thus obtained from
  $D$ so that $H_{i,:}\approx
  P(\cdot|S_i,s-1,\lambda)$. Those conditional distributions
  satisfy the following equality for any $V^{(s-1)}$:
  \[
  P(v_1 V^{(s-1)}|S_i,s,\lambda)=\sum_{j=1}^N a_{i,j}(1)
  P(V^{(s-1)}|S_j,s-1,\lambda).
  \]
  Therefore
  $
  P(v_1 \cdot |S_i,s,\lambda)=\sum_{j=1}^N a_{i,j}(1) P(\cdot|S_j,s-1,\lambda)
  $
  so we can obtain the unknown values $a_{i,j}(1)$ by solving the
  following systems of linear equations:
  \[
  D_{i,1:M^{s-1}}=A_{i,:}(1)*H,\ i=1,2,\ldots, N
  \]
  where $A_{i,:}(1)=[a_{i,1}(1)\ a_{i,2}(1)\cdots
  a_{i,N}(1)]$. Compactly $D_{:,1:M^{s-1}}=A(1)*H$. As in step $2$,
  because of the finiteness of $T$ and working with bounded arithmetic
  precision we need to content ourselves with a solution that
  minimizes some distance (for example, the $L_1$ norm) between $D_{i,1:M^{s-1}}$ and
  $A_{i,:}(1)*H$, for all $i$.  We have formulated these problems
  as linear programming problems using the $L_1$ norm.

\item Output estimated HMM $\lambda'=\{A(k)\ |\
  k=1,2,\ldots,N\}$.

\end{enumerate}
This algorithm can be iterated using the estimated $\lambda'$ and
formula~(\ref{pV}) to compute new matrices $C_0'$ and $D_0'$, and
then restarting from step $3$ above with matrices $C_0'$ and
$D_0'$ as initial factors in the approximate NMF. In particular:

\begin{enumerate}

\item[6)] Compute $D_0'=[P(j|S_i,s,\lambda')]_{i,j}$
  using formula~(\ref{pV}).

\item[7)] Compute $C_0'$ by solving the linear programming problem
$F^{p,s} = C_0'*D_0'$, for a row stochastic $C_0'$.

\item[8)] Set $C_0:=C_0'$ and $D_0:=D_0'$.

\item[9)] goto 3).

\end{enumerate}

Another possibility for step 7) above is to compute $C_0'$ using
formulae (2) and (3) and then use the resulting $C_0'$ and $D_0'$
as initial guesses for the NMF algorithm.  We have tried this
variant but it does not produce significantly different final
results.

\section{Numerical examples}

We call an HMM ``Deterministic'' (DHMM) if for each state there
exists at most one outgoing transition labeled with the same
observable. We demonstrate our method on a DHMM, on an HMM that
can be transformed into an equivalent DHMM and also on an HMM for
which such a transformation does not exist. We finally discuss an
example that illustrates the situation when $rank\ <\ prank$.

It is important to note that the significant metric for learning an
HMM is {\em not} the extent to which the transition probabilities are
accurately learned but the extent to which the observation statistics
are learned.  This is a consequence of the fact that HMM's with
different transition probabilities and different numbers of states can
produce observations sequences with the same statistics so that
learning a specific transition probability characterization is not a
well-posed problem unless additional constraints to the learning problem are
imposed~\cite{ito:1992}.

In our examples we measure the accuracy of our estimates by computing
the I-divergence rate of the finite dimensional distributions
associated with the observation process of the original model from
those associated with the observation process of the estimated
model. Formally, each HMM $\lambda$ induces a
family of finite dimensional distributions
\[
P_n(y_1^n) = \sum_{i=1}^N \pi_i P(y_1^n|x_1=S_i,\lambda)
\]
on sequences of observations of length $n$, where $\pi$ is the
stationary distribution of the underlying state process. Let $\lambda$
and $\lambda'$ be two HMM's with $P_n$ and $Q_n$ their respective
induced finite dimensional distributions. The I-divergence rate of
$\lambda$ from $\lambda'$ is defined as
\[
\overline{D}_{ID}=\lim_{n\rightarrow\infty} \frac{1}{n}D_{ID}(P_n||Q_n)\ 
\]
when the limit exists~\cite{finesso:2008}.

\subsection{A DHMM Example}

Consider the stochastic process described by model
$\lambda_1 = (\{ A(0), A(1)\},\Gamma=[0\ 1])$ with
\[
A(0)=
\left [
\begin{array}{cc}
   0.5 &      0 \\
     0 &      0
\end{array}
\right ]\hspace{0.5cm}\mbox{and}\hspace{0.5cm}
A(1)=
\left [
\begin{array}{cc}
     0 &    0.5 \\
     1 &      0
\end{array}
\right ]\ .
\]
This is sometimes referred to as the ``Even
Process''~\cite{SSC:2002,weiss:1973}. We simulated this process
and produced a sequence of $T=1000$ observations.  Then we ran our
algorithm with $p=2$ and $s=3$:

\begin{enumerate}

\item Build $F^{2,3}$ from data ${\cal O}$:
\[
F=\left [
\begin{array}{cccccccc}
  0.14 &   0.13 &      0 &   0.26 &      0 &      0 &   0.22 &   0.26 \\
     0 &      0 &      0 &      0 &   0.25 &   0.24 &      0 &    0.5 \\
  0.13 &   0.14 &      0 &   0.23 &      0 &      0 &   0.26 &   0.24 \\
 0.08 &  0.07 &      0 &   0.17 &  0.08 &  0.08 &   0.17 &   0.33
\end{array}
\right ]
\]

\item Estimate $N={\rm prank}(F)$. Analyze singular values of $F$:
\[ \left [ \begin{array}{cccc}
  0.88 &   0.48 &  0.033 &  0.011
\end{array} \right ]\ . \]
This suggests ${\rm rank}(F)={\rm prank}(F)=2$.

\item Estimate distributions $P(\cdot|S_1,3,\lambda_1)$ and $P(\cdot|S_2,3,\lambda_1)$ by solving $\arg\min_{C,D} D_{ID}(F||C*D)$:
\[
C =
\left [
\begin{array}{cc}
  0.02 &   0.98 \\
     1 &      0 \\
     0 &      1 \\
  0.34 &   0.66
\end{array}
\right ]\ ,
\]
\[
\hspace{-1cm}D=\left [
\begin{array}{cccccccc}
     0 &      0 &      0 & 0.0 &   0.25 &   0.24 & 0.0 &    0.5 \\
  0.13 &   0.13 &      0 &   0.25 &      0 &      0 &   0.25 &   0.24
\end{array}
\right ]
\]

\item Estimate matrices $A(0)$ and $A(1)$:
\[
\hspace{-0.6cm}\tilde{A}(0)=
\left [
\begin{array}{cc}
2.2e-18 &      0 \\
6.9e-18 &   0.51
\end{array}
\right ],\tilde{A}(1)=
\left [
\begin{array}{cc}
0.0077 &   0.99 \\
  0.49 &      0
\end{array}
\right ]\ .
\]

\end{enumerate}
After a second iteration of the algorithm the reconstructed matrices
become:
\[
\hat{A}(0)=
\left [
\begin{array}{cc}
     0 &      0 \\
5.6e-17 &   0.51
\end{array}
\right ]\ \hat{A}(1)=
\left [
\begin{array}{cc}
0.0077 &   0.99 \\
  0.49 &      0
\end{array}
\right ]\ .
\]
The reconstructed model is essentially identical to the original one
except for state reordering. This result is competitive with
  existing techniques specific for the machine learning of DHMMs. For
  example, Shalizi et al~\cite{shalizi:2004,SSC:2002} demonstrated
  their Causal-State Splitting Reconstruction (CSSR)
  $\epsilon$-machine reconstruction algorithm on the same Even Process
  obtaining comparably accurate models.

\subsection{An HMM that has an equivalent DHMM}

Consider the model $\lambda_2 = (\{ A(0), A(1)\},\Gamma=[0\ 1])$
with
\[
A(0)=
\left [
\begin{array}{cc}
  0.67 &   0.33 \\
  0 &   0
\end{array}
\right ]\hspace{0.5cm}\mbox{and}\hspace{0.5cm}
A(1)=
\left [
\begin{array}{cc}
  0 &   0 \\
  1 &   0
\end{array}
\right ]\ .
\]
We simulated this process and produced a sequence of $T=10000$
observations.  Then we ran our algorithm with $p=2$ and $s=3$:

\begin{enumerate}

\item Build $F^{2,3}$ from data ${\cal O}$:
\[
F =
\left [
\begin{array}{cccccccc}
  0.31 &   0.14 &   0.22 &   0 &   0.22 &   0.11 &   0 &   0 \\
  0.44 &   0.23 &   0.33 &   0 &   0.00 &   0    &   0 &   0 \\
  0.29 &   0.15 &   0.23 &   0 &   0.22 &   0.11 &   0 &   0 \\
  0    &   0    &   0    &   0 &   0    &   0    &   0 &   0
\end{array}
\right ]
\]
%\[
%\left .
%\input{PFA2_dist0-b}
%\right ]\ .
%\]

%\input{NewPFA2_svd}

\item Estimate $N$. Analyze singular values of $F$:
\[ \left [ \begin{array}{cccc}
  0.86 &   0.24 &   0.02 &   0
\end{array} \right ] \]
% $[0.88645\ 0.23274$ $\ 0.014591\ 0]$
to estimate $N$. This suggests again $N=2$.

\item Estimate distributions $P(\cdot|S_1,3,\lambda_2)$ and $P(\cdot|S_2,3,\lambda_2)$. Solve $\arg\min_{C,D} D_{ID}(F||C*D)$:
\[
C =
\left [
\begin{array}{cc}
  0 &   1 \\
  1 &   0 \\
  0 &   1 \\
  0 &   0
\end{array}
\right ]\ ,
\]
\[
D=\left [
\begin{array}{cccccccc}
  0.44 &   0.23 &   0.33 &   0 &   0 &   0 &   0 &   0 \\
  0.30 &   0.15 &   0.22 &   0 &   0.22 &   0.11 &   0 &   0
\end{array}
\right ]
\]
%\[
%\left .
%\input{PFA2_state-b}
%\right ]\ .
%\]

%\input{NewPFA2_newpapprox1}
%\input{NewPFA2_newpapprox2}

\item Reconstruct matrices $\tilde{\lambda}_2=\{\tilde{A}(0),\tilde{A}(1)\}$:
\[
\hspace{-0.2cm}\tilde{A}(0)=
\left [
\begin{array}{cc}
  0.0033 &   0.9967 \\
  0 &   0.6691
\end{array}
\right ],
\tilde{A}(1)=
\left [
\begin{array}{cc}
  0 &   0 \\
  0.3309 &   0
\end{array}
\right ]\ .
\]

\end{enumerate}
After a second iteration of the algorithm the reconstructed model
becomes $\hat{\lambda}_2=\{\hat{A}(0),\hat{A}(1)\}$:
\[
\hat{A}(0)=
\left [
\begin{array}{cc}
  0.0039 &   0.996 \\
  0 &   0.6689
\end{array}
\right ], \hat{A}(1)=
\left [
\begin{array}{cc}
  0 &   0 \\
  0.3311 &   0
\end{array}
\right ]\ .
\]
These computed transition probabilities are different enough from the
transition probabilities of the original HMM used to generate the data
but the statistics of the observation sequences are very
close. Figure~\ref{kl3} shows the accuracy of these estimates in terms
of the I-divergence rate of the original model from the estimated
ones.  We computed $D_{ID}(P_n||Q_n)/n$ for $n=1,2,\ldots, 15$, with
$P_n$ being the finite dimensional probability distributions over
sequences of observations of length $n$ emitted by model $\lambda_2$
and $Q_n$ those emitted by the estimates $\tilde{\lambda}_2$ and
$\hat{\lambda}_2$, in stationary conditions (the dotted curve refers
to $\tilde{\lambda}_2$). We can observe that this quantity, the
divergence rate of $P_n$ from $Q_n$, stabilizes to a very small value
(smaller than $2.5\cdot 10^{-5}$) as expected.

In fact, this example is equivalent to a DHMM model as the reader
can readily check independently.

\subsection{An HMM that has no equivalent finite state DHMM}

\begin{figure}[!t]
\centering
\includegraphics[width=3.5in]{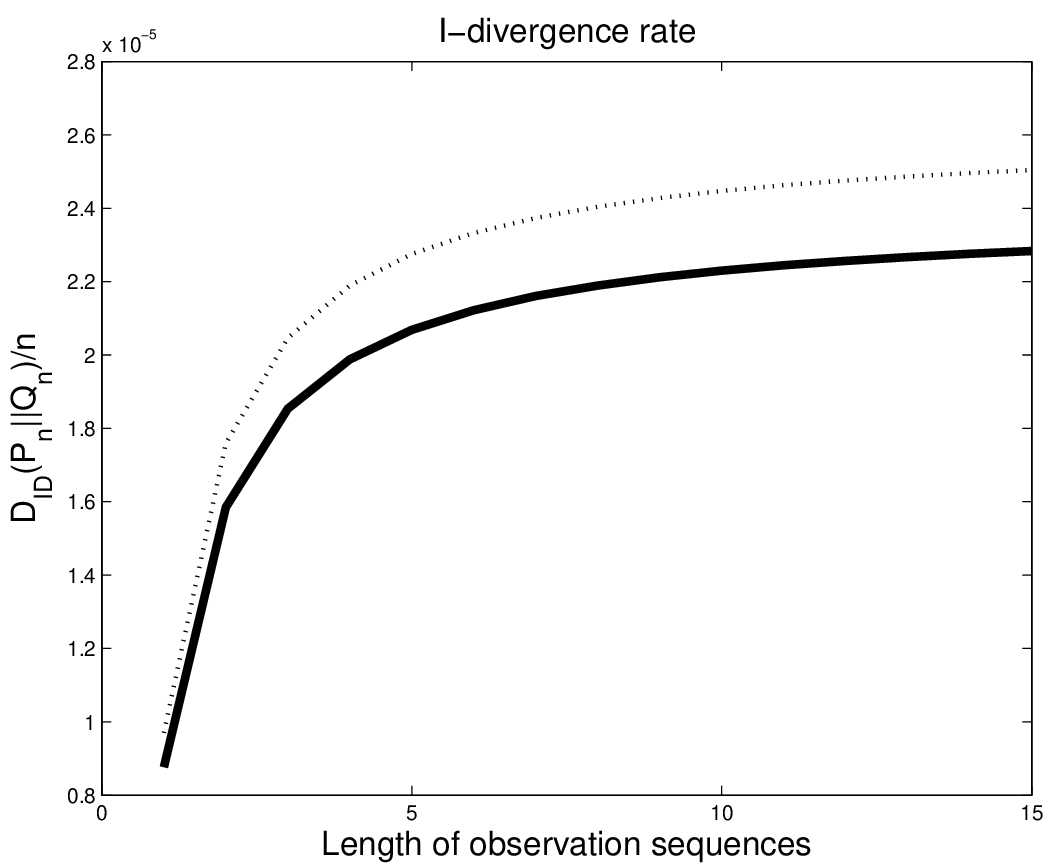}\\
\includegraphics[width=3.5in]{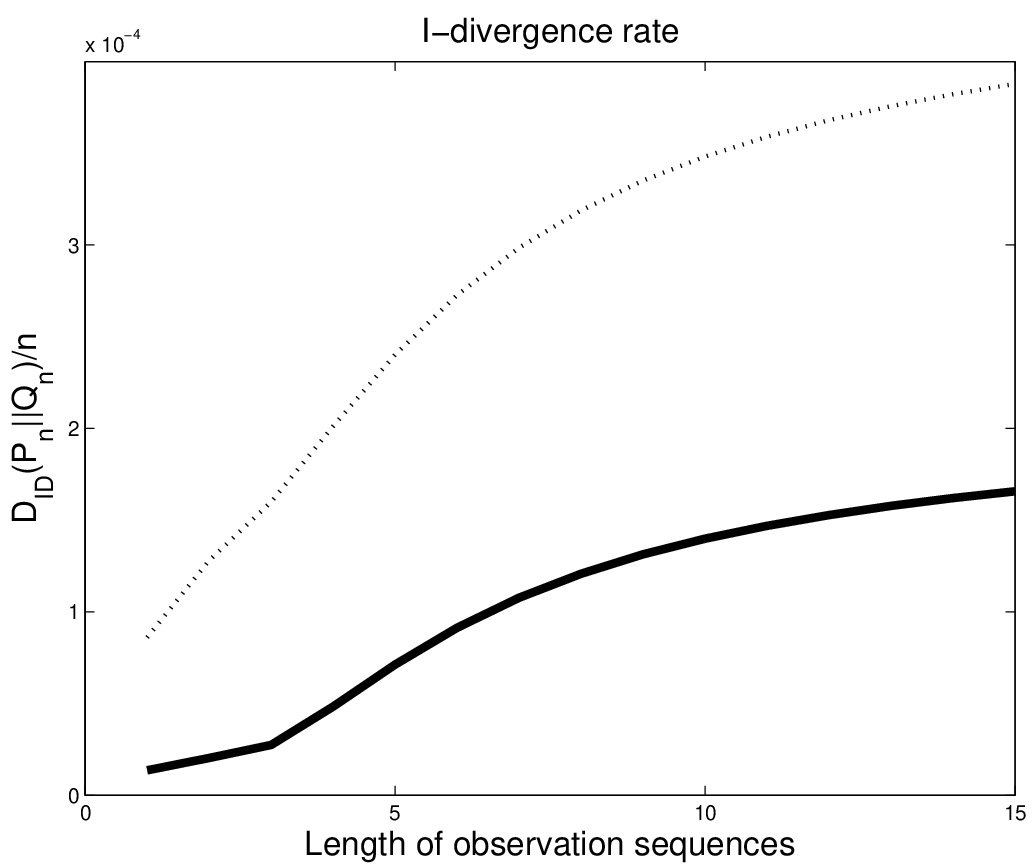}
\caption{Accuracy of HMM's $\tilde{\lambda}_2,\hat{\lambda}_2$ (top)
  and $\tilde{\lambda}_3,\hat{\lambda}_3$ (bottom). Here $P_n$ is the
  distribution over sequences of observations of fixed length $n$
  induced by the original model whereas $Q_n$ refers to the estimated
  models. The sequence $D_{ID}(P_n||Q_n)/n$ is calculated for
  increasing values of $n$. The dotted curves refer to
  $\tilde{\lambda}_2$ (top) and $\tilde{\lambda}_3$ (bottom).}
\label{kl3}
\end{figure}

Consider the model $\lambda_3 = (\{ A(0), A(1)\},\Gamma=[1\ 0\ 0])$
with
\[
A(0)=
\left [
\begin{array}{ccc}
  0.5 &   0.5 &   0 \\
  0 &   0.5 &   0 \\
  0.5 &   0.5 &   0
\end{array}
\right ]\hspace{0.5cm}\mbox{and}\hspace{0.5cm}
A(1)=
\left [
\begin{array}{ccc}
  0 &   0 &   0 \\
  0 &   0 &   0.5 \\
  0 &   0 &   0
\end{array}
\right ]\ .
\]
We simulated this process and produced a sequence of $T=10000$
observations.  Then we ran our algorithm with $p=4$ and $s=5$.
After the first iteration we obtain
$\tilde{\lambda}_3$:
\[
\tilde{A}(0)=
\left [
\begin{array}{ccc}
  0 &   0.2 &   0.8 \\
  0 &   0.35 &   0 \\
  0 &   0.4 &   0.6
\end{array}
\right ], \tilde{A}(1)=
\left [
\begin{array}{ccc}
  0 &   0 &   0 \\
  0.1 &   0 &   0.56 \\
  0 &   0 &   0
\end{array}
\right ]\ .
\]
After the second iteration we obtain $\hat{\lambda}_3$:
\[
\hat{A}(0)=
\left [
\begin{array}{ccc}
  0 &   0.2 &   0.8 \\
  0 &   0.36 &   0 \\
  0 &   0.41 &   0.59
\end{array}
\right ], \hat{A}(1)=
\left [
\begin{array}{ccc}
  0 &   0 &   0 \\
  0.09 &   0 &   0.56 \\
  0 &   0 &   0
\end{array}
\right ]\ .
\]
As before, Figure~\ref{kl3} (bottom) shows the accuracy of these
estimates in terms of the I-divergence rate of the original model from
the estimated ones.

Observe that this HMM cannot be transformed into an equivalent
deterministic HMM~\cite{Shalizi:1999}.

\subsection{Discussion of Rank vs Prank}

We first provide an example of a stochastic matrix whose ${\rm
prank}$ differs from its ${\rm rank}$ but that matrix does not
represent the statistics of any HMM.
\[
F=\frac{1}{16}\cdot\left [
  \begin{array}{cccc}
    1 & 1 & 0 & 0\\
    1 & 0 & 1 & 0\\
    0 & 1 & 0 & 1\\
    0 & 0 & 1 & 1
  \end{array}
\right ]\otimes [1\ 1\ 1\ 1\ 1\ 1\ 1\ 1]\ ,
\]
where $\otimes$ is the Kronecker product. We can verify that ${\rm
  rank}(F)=3$ whereas ${\rm prank}(F)=4$~\cite{jcohen:1993}. Moreover
$F=CD$ exactly with
\[
C = \left [
  \begin{array}{cccc}
    0.5    &     0 &   0.5   &      0\\
    0.5 &   0.5  &  0  &       0\\
    0   &      0  &  0.5  &  0.5\\
    0  &  0.5   &      0  &  0.5
  \end{array}
\right ]
\]
and
\[
D = \frac{1}{8}\left [
  \begin{array}{cccc}
    1 & 0 & 0 & 0 \\
    0 & 0 & 1 & 0\\
    0 & 1 & 0 & 0\\
    0 & 0 & 0 & 1
  \end{array}
\right ]\otimes [1\ 1\ 1\ 1\ 1\ 1\ 1\ 1]\ .
\]
Assume that $F$ was obtained from a typical sequence of
observations emitted by an HMM $\lambda$ with $4$ states so that
$F^{2,5}=F=CD$. Then it must be that
$C=[P(S_j|i,2,\lambda)]_{i,j}$ and $D=[P(k|S_j,5,\lambda)]_{j,k}$.
Consider the following model $\lambda=\{A(1),A(2)\}$ with
$$
A(0) = \left [
  \begin{array}{cccc}
    0.5 &   \hspace{-0.2cm}     0  & \hspace{-0.2cm} 0.5   & \hspace{-0.2cm}     0\\
    0     &  \hspace{-0.2cm}  0  &  \hspace{-0.2cm}     0 &  \hspace{-0.2cm}      0\\
    0 &  \hspace{-0.2cm} 0.5 &   \hspace{-0.2cm}      0  & \hspace{-0.2cm} 0.5\\
    0  &  \hspace{-0.2cm} 0      & \hspace{-0.2cm}  0  & \hspace{-0.2cm} 0
  \end{array} \right ],
A(1) = \left [
  \begin{array}{cccc}
    0 &  \hspace{-0.2cm}      0  & \hspace{-0.2cm} 0   &  \hspace{-0.2cm}    0\\
    0.5     &  \hspace{-0.2cm}  0  &  \hspace{-0.2cm}     0.5 &   \hspace{-0.2cm}     0\\
    0 & \hspace{-0.2cm}  0 &   \hspace{-0.2cm}      0  & \hspace{-0.2cm} 0\\
    0  &  \hspace{-0.2cm} 0.5      &  \hspace{-0.2cm} 0  &  \hspace{-0.2cm} 0.5
    \end{array} \right ]\ .  $$
  One can verify that $\lambda$ is the only four-state model such that
  $D=[P(k|S_j,5,\lambda)]_{j,k}$. In fact observe that the system
  of equations defining $\lambda$ in stage $4$ of the algorithm
  admits, in this case, only one solution. Nevertheless, using
  formula~(\ref{pS}):
  \[
  [P(S_j|i,2,\lambda)]_{i,j} = (1/4)*
  \left [
    \begin{array}{cc}
      1 & 1\\
      1 & 1
    \end{array}\right ]\otimes
  \left [ \begin{array}{cc}
      1 & 1\\
      1 & 1
    \end{array}\right ]\neq C\ .
  \]
  Consequently no HMM can generate $F$.

\subsubsection{An example of prank $>$ rank for an exact HMM model}

The following four-state model $\lambda$ is an  example of an HMM
whose induced $F^{2,5}$ matrix has rank $3$ but positive rank
$prank=4$:
\[
A(0)=
\left [
\begin{array}{cccc}
   0.5 &      0 &      0 &      0 \\
     0 &      0 &      0 &      0 \\
     0 &      0 &      0 &      0 \\
     0 &      0 &      0 &      0
\end{array}
\right ],~ A(1)= \left [
\begin{array}{cccc}
     0 &    0.5 &      0 &      0 \\
     0 &      0 &      1 &      0 \\
     0 &      0 &      0 &      1 \\
     1 &      0 &      0 &      0
\end{array}
\right ].
\]
To verify the claim we computed factors
$C=[P(S_j|i,\lambda)]_{i,j}$ and $D=[P(j|S_i,5,\lambda)]_{i,j}$,
for $N=4$, using formulae~(\ref{pV}) and~(\ref{pS}) and then
obtained $F^{2,5}=C*D$. Then we verified numerically that ${\rm
rank}(F^{2,5})=3$. Finally, we applied Lemma 2.4
in~\cite{jcohen:1993} to confirm that ${\rm
  prank}(F^{2,5})=4$. We also verified the character of this model by directly
applying  our algorithm to it in order to obtain $F^{2,5}$
empirically (for $T=10000$).  An analysis of the singular values
of $F^{2,5}$, namely $[0.8530\ 0.4825\ 0.1799\ 0.0114]$,
%$[0.3285\ 0.0506\ 0.0316\ 0.0012]$
demonstrates the difficulty of this case. The fourth singular
value is nonzero due to the finiteness of $T$. Consequently it is
difficult to determine whether $N=3$ or $N=4$.

\section{Open Questions and Future Work}

A  crucial issue is the estimation of $N$, the size of the
smallest HMM that generates the stream of data. Under ideal
conditions, ($T\rightarrow\infty$), we have seen that $N={\rm
prank}(F^{p,s})$. However,  filtering out ``noise'' from the
empirical matrix $F^{p,s}$ in order to have an accurate estimate
of the positive rank is an open challenge. Observe that a
spectral analysis of $F$ may, in general, produce only a lower
bound to $N$.

A second important issue in our methodology concerns  the
computation of the approximate NMF. Existing methods are
suboptimal due to the presence of local optima. This problem
affects the accuracy of the produced estimate at each iteration of
our algorithm. Consequently it is important to investigate
convergence properties when stages $3-5$ of the algorithm are
iterated with new initial factors $C_0',D_0'$ to seed the
approximate NMF, using $C_0'$ and $D_0'$ as computed according to
steps $6-8$, from model $\lambda'$ that was estimated in the
preceding step.

A third question concerns with properties of $F^{p,s}$ as
$s\rightarrow\infty$. In other words, can the Asymptotic
Equipartition Property be applied to distributions
$P(\cdot|S_i,s,\lambda)$ so that the distribution on the
``typical'' finite suffixes is uniform and the rest of the
distribution is zero?

\section{Conclusion}
We have presented a new algorithm for learning an HMM from
observations of the HMM's output.  The algorithm is structurally
different from traditional Baum-Welch based approaches
\cite{Rabiner:1989,willsky:2008,blei04variational,teh03hierarchical}.
It  is related to but different from recent approaches in
stochastic systems realization \cite{finesso:2008}. We believe
this method opens a new line of algorithm development for learning
HMM's and has the advantage of a estimating the HMM order from
spectral properties of the high order correlation statistics of
the observation sequence.  The algorithm effectively compresses
data by summarizing it into a statistical matrix.  Options for
recursively computing the steps of the algorithm to achieve {\em
on-line} algorithms will be explored.  Additionally, sparse matrix
algorithms can be explored for space and time efficiency when the
underlying matrices are large and sparse.

\section*{Acknowledgments}

This work results from research programs at the Thayer School of
Engineering, Dartmouth College, and at the Department of Computer
Science at the California State University, Los Angeles, supported by
the AFOSR Grant FA9550-07-1-0421 and by the NSF Grant HRD-0932421. The
Dartmouth effort was also supported by U.S. Department of Homeland
Security Grant Award Number 2006-CS-001- 000001 and DARPA Contract
HR001-06-1-0033.

The authors also sincerely thank Dr. Robert Savell for pointing
out important related work to us.


\ifCLASSOPTIONcaptionsoff
  \newpage
\fi



\bibliographystyle{IEEEtran}
%\bibliography{IEEEabrv,hmmLearning,hmmDatabase,reviewersrefs}

%\begin{thebibliography}{1}

%\bibitem{HMM}
%HMM papers, etc

%\end{thebibliography}

%\begin{biography}[{\includegraphics[width=1in,height=1.25in,clip,keepaspectratio]{mshell}}]{Michael Shell}

%\begin{IEEEbiography}{George Cybenko}...\end{IEEEbiography}

\begin{biography}
%[{\includegraphics[width=1in,height=1.25in,clip,keepaspectratio]{Hartford-20100903-00004}}]
{George Cybenko}
  is the Dorothy and Walter Gramm Professor of Engineering at the
  Thayer School of Engineering at Dartmouth College.  Prior to joining
  Dartmouth, he was Professor of Electrical and Computer Engineering
  at the University of Illinois at Urbana-Champaign. His current
  research interests are in machine learning, signal processing and
  computer security. Cybenko was founding Editor-in-Chief of IEEE
  Computing in Science and Engineering and IEEE Security \&
  Privacy. He is presently First Vice-President of the IEEE Computer
  Society. He earned his B.Sc. (Toronto) and Ph.D (Princeton) degrees
  in mathematics and is a Fellow of the IEEE.
\end{biography}

\begin{biography}
%[{\includegraphics[width=1in,height=1.25in,clip,keepaspectratio]{crespifoto2}}]
{Valentino Crespi}
  received his Laurea Degree and his Ph.D. Degree in Computer Science
  from the University of Milan, Italy, in July 1992 and July 1997,
  respectively. From September 1998 to August 2000 he was an Assistant
  Professor of Computer Science at the Eastern Mediterranean
  University, Famagusta, North Cyprus and from September 2000 to
  August 2003 he worked at Dartmouth College, Hanover, NH, as a
  Research Faculty. Since September 2003 he has been on faculty at the
  Department of Computer Science, California State University, Los
  Angeles, currently in the capacity of Associate Professor. His
  research interests include Distributed Computing, Tracking Systems,
  UAV Surveillance, Sensor Networks, Information and Communication
  Theory, Complexity Theory and Combinatorial Optimization. At
  Dartmouth College he developed the TASK project and consulted to
  the Process Query Systems project, directed by Prof. George
  Cybenko. At CSULA he has been teaching lower division, upper
  division and master courses on Algorithms, Data Structures, Java
  Programming, Compilers, Theory of Computation and Computational
  Learning of Languages and Stochastic Processes. During his
  professional activity Dr Crespi has published a number of papers in
  prestigious journals and conferences of Applied Mathematics,
  Computer Science and Engineering. Moreover Dr Crespi is currently a
  member of the ACM and of the IEEE.
\end{biography}

%\newpage


\vfill


\end{document}